# Explainable Misinformation Detection across Multiple Social Media Platforms

Gargi Joshi[1], Ananya Srivastava[1], Bhargav Yagnik[1], Mohammed Hasan[1], Zainuddin Saiyed[1], Lubna A Gabralla[3], Ajith Abraham[4], Rahee Walambe[1,2*] and Ketan Kotecha[1,2*].

[1] Symbiosis Institute of Technology, Symbiosis International (Deemed University), Pune, India

[2] Symbiosis Centre for Applied Artificial Intelligence, Symbiosis International (Deemed University), Pune, India

[3] Department of Computer Science and Information Technology, Princess Nourah Bint Abdulrahman University, Saudi Arabia

[4] Machine Intelligence Research Laboratories, Auburn, WA 98071, USA

*Corresponding Author: rahee.walambe@sitpune.edu.in (R.Walambe) and director@sitpune.edu.in (K. Kotecha)

"This research was funded by the Princess Nourah bint Abdulrahman University Researchers Supporting Project number (PNURSP2022R178). The research is supported through the Scheme for Promotion of Research Collaboration (SPARC) of the Ministry of Human Resources Development (MHRD), Government of India (Sanction ID- P571).

**ABSTRACT** Web information Processing (WIP) has had an enormous impact on modern society since a huge percentage of the population relies on the internet to acquire information. Social Media platforms provide a channel for disseminating information and a breeding ground for spreading misinformation, creating confusion and fear among the population. One of the techniques for the detection of misinformation is machine learning-based models. However, due to the availability of multiple social media platforms, it has become a tedious job to develop and train AI-based models individually. Despite multiple efforts to develop machine learning-based methods for identifying misinformation, there has been very limited work on developing an explainable generalized detector capable of robust detection and generating explanations beyond black-box outcomes. It is essential to know the reasoning behind the outcomes to make the detector trustworthy. Hence employing the explainable AI techniques is of utmost importance. In this work, the integration of two machine learning approaches, namely domain adaptation and explainable AI, is proposed to address these two issues of generalized detection and explainability. Firstly the Domain Adversarial Neural Network (DANN) develops a generalized misinformation detector across multiple social media platforms. DANN is employed to generate the classification results for test domains with relevant but unseen data. The DANN-based model, a traditional black-box model, cannot justify and explain its outcome, i.e., the labels for the target domain. Hence a Local Interpretable Model-Agnostic Explanations (LIME) explainable AI model is applied to explain the outcome of the DANN mode. To demonstrate these two approaches and their integration for effective explainable generalized detection, COVID-19 misinformation is considered a case study. We experimented with two datasets and compared results with and without DANN implementation. It is observed that using DANN significantly improves the F1 score of classification and increases the accuracy and AUC. The results obtained show that the proposed framework performs well in the case of domain shift and can learn domain-invariant features while explaining the target labels with LIME implementation. This can enable trustworthy information processing and extraction to combat misinformation effectively.

## I. INTRODUCTION

Information and data are primarily stored on various social networks and internet platforms, and web information processing offers opportunities for modification and extraction of the data[1]. Web Information Processing (WIP) manipulates data available from various internet/web sources to produce useful information. One of the specific aspects of web information processing includes extracting and using the information available on social media, including the internet and social networking websites. A considerable percentage of the population relies on the internet for acquiring information. Social Networking platforms are growing each day, with 4.2 billion users in January 2021 and a 13.2% increase in 2020 alone. The rate has doubled compared to 2019-2020, which was only 7.2% [2]. Due to this spike, the enormous amount of misinformation communicated on social media platforms poses an unprecedented challenge causing significant harm and being deleterious to many people worldwide. Misinformation can be any inadvertent premise not backed by facts or scientific data and leads to misconceptions. This is highly important in the case of healthcare-related misinformation, which can lead to fatal consequences. Furthermore, the circulation of misinformation during a health crisis induces trepidation among the population. For example, detail about COVID-19 from these social media posts and news sources can include opinionated "natural remedies," the origin of COVID-19, or about the vaccines and their side effects cascade hesitancy in getting vaccinated even if vaccines are available.

The type of language and ideas on a specific topic portrayed by the users from one social media platform to another significantly differs, starting from vocabulary, grammar, etc.[3]. This also differs from news data sources. Annotating data from multiple information pools can be tedious, time-consuming, and costly, which is a major reason many researchers usually focus on one specific data source when classifying misinformation/fake news. In the case of a pandemic, dependency on a particular source drastically affects the robustness of a classifying model to be applied on other social networks when the spread of misinformation across all platforms is gradually increasing and equally important.

Creating an adaptive model over diverse media platforms can be done effectively by learning information from one domain and utilizing it to learn in another domain. Domain Adaptation [4] deals with such generalization beyond training distribution and comes under the purview of out-of-distribution generalization[5]. This is accomplished using Domain Adversaries to learn domain invariant feature representations. DANN applies the gradient reversal layer to make the feature distribution of source and target domains similar. Data distribution differs from one social media platform due to user behavior; data from different platforms constitute different domains.

However, it is important to note that the DANN-based approach is a traditional black-box model, and the outcome, i.e., the reasons for the generated target labels, are not explained. Despite multiple efforts to develop machine learning-based methods for identifying misinformation, very limited work has focused on providing explanations beyond a black-box decision [6]. To develop a trustworthy generalized detector for social media misinformation, explanations must be provided for the target labels. Hence in this work, we employ a DANN and implement it to detect misinformation in coronavirus-related posts across these domains, followed by a Local Interpretable Model-Agnostic Explanations (LIME)[7] framework to generate the reasoning behind the outcomes.

II. RELATED WORK

The WIP consists of multiple facets: information extraction and making it available for internet users. The internet has become a tool for generating and free flow of information worldwide in today's world. Information generates ideas and drives decisions. However, the internet generates false information since no regulatory body moderates the content, especially on social media and WIP, impacting society [8]. For the last few decades, the major concern was air and sound pollution; however, now is the time to worry about information pollution. Information pollution's direct reasons and impacts are difficult to identify, explain, and even more challenging to quantify. In [9], the information disorder phenomenon is examined comprehensively. In many instances, there is no malicious intent to generate and spread misinformation; however, in other cases, it might be just very selfish, e.g., increasing the sale of a certain drug or purely being business-oriented, to spread wrong information to create panic about certain drug or treatment. It is interesting to understand the ethical and moral status of the people involved. When people cannot tell what is credible and what is not and act on that information, poor decisions can be made that can impact our lives and our financial wellbeing[10]. We must recognize that communication and information sharing plays a significant role in representing shared beliefs. Since social platforms are designed to express through likes, comments, and shares, all the efforts towards fact-checking and debunking false information are ineffective since the emotional aspect of sharing information is impossible to control. The mining of misinformation in social media spreads uncontrollably and tremendously fast and, in recent times, has been responsible for causing [11] harm to the social fabric of our world. Misinformation can be disinformation, rumors, spam, fake news, etc. The rampant spread of misinformation online is considered one of the ten global risks by the world economic forum [12].

Regarding misinformation, "an era of fake news" is occurring rapidly where misinformation is transmitted speedily on an intentional or unintentional basis. Various formats of non-textual media such as bit-mapped pictures are being used, contributing to the diffusion of misinformation and disinformation. Misinformation affects communities in various ways, for instance, racist hostility and exclusion. Therefore, preventing such emerging behaviors is an important area of research and study. Pizzagate, the Anti-vaccine movement, Russian scientists discovering a cure for homosexuality, etc., are some of the popular fake news items of 2017[13][14], suggesting that curbing social bots may be an effective strategy for mitigating the spread of online misinformation. It is also important to understand that although much of the misinformation is focused on the political domain, medical misinformation has threatened countries worldwide. Research has demonstrated how inaccurate advice from a person who has no medical knowledge is proliferated through hoaxes, tweets [15], online Q&A forums, Pinterest [16], Yahoo, and Google [17]. The context of medical information encompasses the broader aspects of trustworthiness, reliability, dependability, integrity, and reputation of the medical practitioner and the AI developer in the high-risk health and safety domain. In recent times, due to the explosion and wide acceptance of social media reporting, the issue of non-credible news has become extremely relevant. In the context of medical information, this problem is even more serious because it directly relates to the health and well-being of people. Medical misinformation is an obvious reason for concern as the information can be shared without any rigorous review process. The first step toward handling the spread of misinformation is to detect it automatically. Machine learning and artificial intelligence methods are employed for this task, especially for automatic misinformation detection on social media platforms.

*A. DETECTION OF MISINFORMATION USING ARTIFICIAL INTELLIGENCE AND MACHINE LEARNING*

Earlier efforts on misinformation detection go back to the start of the internet revolution Kinchla, and Atkinson [18] have studied the effect of false information on psychophysical judgments. Their experimental results show that false information reduces the probability of a correct response. Various applications, including credibility assessment of microblogs[19], have been reported recently. The credibility assessment algorithms are automated, human-based, and hybrid approaches. Automated approaches include various machine learning approaches. Human-based can be the voting, cognitive, and manual verification approaches. Hybrid approaches combine these. Shao et al. [19] have developed Hoaxy, an open platform for dealing with misinformation. The huge scale of information (data), dynamic nature, and homophily are primary challenges in detecting misinformation. To this end, various researchers have studied and developed methods to see false information. With the flood of information from social websites, it is impossible to check, analyze and vet the potential deception. Various methods starting from simple classifiers to the state of the art attention networks, have been reported in this regard [20]–[24]. However, these approaches are explicitly reported for a single social media platform and information classification. No method can be used across various platforms. This work considers a state-of-the-art paradigm named domain adaption for developing a misinformation detector trained on one source domain and adapted to multiple similar domains.

*B. DOMAIN ADAPTATION*

Domain adaptation is classified as a transductive learning procedure [25] under transfer learning [26]. It is applied in problems where the task stays the same across domains; however, data may not be available in the target domain. Initially, domain adaptation setups considered this problem a supervised approach or a semi-supervised approach with substantial sources and scant data in the target domain. [27][28] proposed solutions to the domain shift problem using such setups. However, with the difficulties in annotated data, unsupervised domain adaptation techniques emerged that alleviate the domain shift problem. Domain adaptation techniques are classified into data-centric, model-centric, and hybrid methods [29]. Data-centric approaches include methods like pseudo labeling [30]–[32] and Data Selection [33], [34]. The model-centric approach focuses on modifications in the architecture that may include the use of pivot-based methods for feature augmentation [26], [35], [36] that have now been developed to learn from neural networks [37], [38] via attention [39], [40]. Feature generalization techniques are capable of learning common hidden features between domains, and such approaches include stacked denoising auto-encoders [41] and marginalized stacked denoising auto-encoders [42], [43], and Dual Representation based auto-encoder [44]. [45] proposed considering features from either source or target using domain separation networks; however, they lacked domain-specific features in the classifier as they were only used in the decoder. [46] Proposed a gradient reversal technique to maximize domain confusion and minimize the task error. DANN-based approaches have widely been applied to applications like sentiment classification [46]–[48], language identification [49], duplicate question detection [50], etc. [51] applied domain adversarial training leveraging the knowledge distillation [52] with an extra loss during adaptation. Hybrid approaches like [48], [53]–[55] combine data-centric and model-centric approaches. Unlike DANN, a loss-centric approach, the feature-centric and data-centric techniques are used in text classification but do not use context-dependency and linguistic information and are multi-shot training procedures [28]. Also, there has been very little research tackling domain differences across social media platforms. In this work, we employ DANN to develop a generic misinformation classifier across multiple platforms. We consider one or more social media platform(s) as the source domain and other multiple platforms as target domains and apply DANN to demonstrate our approach. The most relevant case study of COVID-19 misinformation is considered for the demonstration. However, due to black box nature of the DANN, it is difficult to generate the explanations for the target labels. Hence to that end, the explainable AI is essential.

*C. EXPLAINABLE ARTIFICIAL INTELLIGENCE FOR WEB INFORMATION PROCESSING*

Machine learning models have recently achieved tremendous progress in performance and accuracy. Still, the complex non-intuitive hidden layer processing makes them an opaque black box with a lack of insight into how and why a model generated a certain decision/outcome[56]. This black Box nature results in little or no understanding of the model's internal logic, adversely affecting these models' trust, usability, and adoption in real-world applications[57]. Model interpretability or explainability is of paramount importance when it comes to debugging the model for flaws, ensuring trust,
transparency, accountability, and ethics in the model's outcomes, and complying with the governance such as EU-GDPR, which allows the end-user to seek the right to explain automated algorithmic decisions. XAI improves model performance by performing internal audits and bias detection[58]. The natural language processing (NLP) text models typically deployed for detecting misinformation on Large-scale social media platforms demand accurate introspection and justification of the model underlying Predictions to ensure trust, transparency, and fair decisions from different stakeholders[59]. A prime concern is to generate human-understandable comprehensive model explanations to interpret the model predictions and underlying mechanics [60]. State-of-the-art machine learning models are complex black boxes facing accuracy interpretability trade-offs raising concerns on reliable and reasonable behavior of models in the real world [61]. Reliability on mere performance and accuracy metrics is not enough, and interpretability is a prime metric for establishing trust in the inner working and logic for accurate model predictions and decision-making[62]. Model interpretability can be achieved intrinsically by design faithful and consistent with the model or post hoc, i.e., explaining the predictions after model building without compromising the accuracy interpretability trade-off. The model's scope is local to a specific instance or globally applicable to the entire model [63]. Model explainability techniques are broadly classified into intrinsic built-by design and post hoc explainability techniques that are irrespective of the underlying model and are derived post-model development without impacting the accuracy and design of the model[64]. The most widely used techniques for post hoc explainability in the literature are LIME (Local Interpretable Model Agnostic Explanations) and SHAP (Shapely Additive Explanations), Deep Shap, Deep Lift, and Cxplain). Explainability techniques are crucial in overcoming the black-box nature of classifiers for unimodal and multimodal deep neural nets for vision and language processing [65].LIME is a model agnostic explainability technique applied to any classifier. The model is learned by perturbing the input data samples and understanding how the predictions change based on the change in the inputs. LIME modifies a single data instance by tweaking the individual feature values and observing the effect on the model outcome. This provides insights into why a specific prediction was made or which variable contributed to the prediction. Feature relevance depicts the importance of individual features on the model predictions and the influence of different features over the model outcomes.
To interpret which model captures linguistic knowledge and semantic details and why a certain prediction is made, explanations can be derived for the predictions focusing on perturbed inputs[66]. This approximates the underlying classifier model with a second model learned by perturbing the original instance. This enables one to identify input components that have the most significant influence or impact on predictions. This approach is model agnostic, and it is easier to learn explanations on a locally weighted dataset than approximate a model globally. Local Interpretable Model-Agnostic Explanations (LIME)[67] generates locally faithful explanations, and it learns an interpretable model locally around the specific prediction. LIME provides interpretable data representations for non-expert users representing the presence and absence of faithful and consistent words to the local model without impacting model performance

*D. RESEARCH QUESTIONS*

In summary, the research questions based on the literature survey are:
1. Can we train a classifier on one source domain and use it for testing on a similar but unseen target domain using machine learning?
2. Can we develop an explainable model integrated with the generic detector to explain the target domain labels?

*E. CONTRIBUTIONS*

In summary, the contribution of this work is:

1. Development of a framework for a classifier that can be used across multiple social media platforms effectively for misinformation classification. Such an approach can be helpful when limited data is available for training.
2. Implementation of the DANN mentioned above architecture outperforms the state-of-the-art results on the CoAID dataset.
3. The development of a novel misinformation dataset (MiSoVac) related to COVID-19 vaccination was collected from news platforms and social media sites and demonstrated the proposed method.
4. The XAI LIME-based approach for explaining the target labels invokes more trustworthiness in the developed DANN model for generalized explainable misinformation detection.

*F. NOVELTY OF THE WORK*
1. Development of techniques capable of generalizing on multiple data of a similar domain.
2. Development of economic approach in terms of time, processing, and efficiency without training models on individual platforms.
3. Consideration of domain difference in social media converging to general features accompanied with explainable, trustworthy adoption paradigm addressing domain adaptation and explainability.

The rest of the paper has been organized as follows: the dataset and methods section describes the CoAID and MiSoVac data sets and the methods, especially DANN and LIME. The results and experiments section describes the performed experiments, corresponding results, and evaluation metrics, followed by the discussion section. The article ends with the conclusion section and a discussion on the future scope of the study

## III. DATASET

Misinformation is spread around us in many forms, including newspapers, television, and the internet. However, most misinformation is spread across social media domains like Twitter, Instagram, Facebook, YouTube, Reddit, and WhatsApp. For this research, we focused on developing a generic misinformation detection with a specific topic of COVID-19; we curated the MiSoVac dataset, which includes data from multiple social media platforms related to vaccines. We also considered the CoAID dataset (openly available) to demonstrate the DANN-based approach XAI.

### A. CoAID

COVID-19 heAlthcare mIsinformation Dataset (CoAID)[68] includes 4251 news articles, 296000 related user engagements, and 926 social platform posts fact-checked by verified fact-checking sites. The data set is collected from articles published between December 2019 to September 1, 2020. Topics like COVID-19, coronavirus, pneumonia, flu, lockdown, stay home, quarantine, and ventilator were covered for the data set

### B. MiSoVac

This dataset was explicitly developed to focus on the case study of COVID-19 vaccine-related misinformation. Therefore, we collected the COVID-19 vaccine-related misinformation data (MiSoVac) from social media sites like Twitter, Instagram, YouTube, and Reddit from November 2020 to February 2021. The details of the same are summarized below

TABLE I
SAMPLES FROM THE MISOVAC DATASET

| Misinformation type (True) | Misinformation type (False) | Misinformation type (None) |
|---|---|---|
| Twitter | | |
| @WHO Solidarity Trials are also underway in many countries. Once these projects are complete controlling #COVID19 effectively will be easier due to the availability of proper medicines, vaccines, etc., as did to control various infectious diseases worldwide. | An article claims that "Bill Gates' vaccine" would modify human DNA. | - |
| Reddit | | |
| The antibodies decrease by a factor of six for SA. Does anyone know what this means/does | If you've had the initial variant, are the antibodies no longer effective against new variants? | And if it's suitable for Moderna, it's likely good for Pfizer. The good news today. See you in the next variant story |
| News | | |
| People who are pregnant, breastfeeding, or want to become pregnant can get vaccinated against COVID-19. But they should talk to their medical provider. | COVID vaccines made with RNA are not vaccines, but a gene therapy that could turn us into transgenic beings or cause us diseases | |
| YouTube | | |
| All vaccines are experimental. With no long-term effects established, one should understand that the Pharmaceutical company is not liable if severe complications or death occurs. It's a take at your risk option. … | The vaccine has been weaponized! | Great video, Vox! The animation, the music, and the narration were all well-made. |
| **Instagram** | | |
| @jamesr.french: I do not think so, but I know that the symptoms may last for a month or two after being infected with COVID 19 virus. | 😂😂😂 biggest scam in history. No flu deaths, no pneumonia deaths, no influenza deaths... just "COVID." 😂😂😂 wtfe! It takes 5 to 10 years to come up with a vaccine before it ever goes to market, yet all of a sudden, in just 5 to 7 months, we have a life-saving vaccine. 😊 the same vaccine that has killed people…. | Sick and tired of this already, till when most we continue living like this |

TABLE II
MiSoVac DATA DISTRIBUTION OF VARIOUS SOCIAL MEDIA PLATFORMS

| Source | No. of Samples Labelled True (contain correct information) | No. of Samples Labelled False (contain misinformation) | No. of Samples labelled None. (contain unimportant information) | Total |
|---|---|---|---|---|
| Twitter | 182 | 346 | 0 | 528 |
| Reddit | 39 | 7 | 322 | 368 |
| News | 149 | 185 | 0 | 334 |
| YouTube | 4 | 10 | 484 | 498 |

| Instagram | 30 | 59 | 2336 | 2425 |

## IV. METHODS

### A PREPROCESSING

The information on social media is usually easygoing and casual, leading to a decrease in the ability of a language model to comprehend the corpus; consequently, performing broad preprocessing on the information became necessary[69]. (Fig. 1) addresses the diagrammatic flow for preprocessing. The underlying pipeline included the expansion of contractions and truncations into their standard form. Basic pronouns, conjunctions, articles, and relational words in English vocabulary usually add no logical importance to a sentence and are disregarded by search engines; hence were eliminated from the corpus. Expulsion of URLs, hashtags, mentions, and punctuations was likewise completed as a part of preprocessing. The emojis were supplanted with the content indicating its importance.

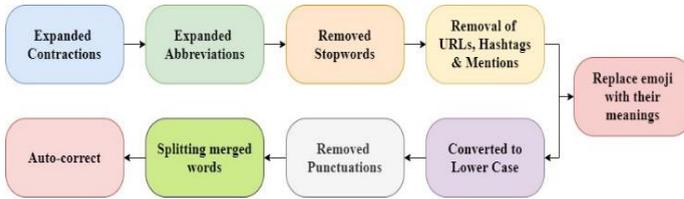

**FIGURE.1.** Preprocessing pipeline and an example

### B DATA AUGMENTATION

The target labels in the MiSoVac dataset were imbalanced, causing the model incapable of generalizing on both the classes, due to which augmentation had to be performed. Table 3. Shows an example of a sentence augmented from the MiSoVac data set. Following operations were performed using [70]:
1. Augmenting words by feeding neighboring words to the BERT language model leverages contextual word embedding.
2. Translation of text into other languages and then translating back to English sentences. The following languages were used in this approach: French, Japanese, German, and Urdu.

TABLE III
EXAMPLE OF DATA AUGMENTATION

| Data | Augmented Data |
|---|---|
| Last night I closed my eyes and saw that people in the hospital room had been vaccinated with cod. | I closed my eyes last night. I've seen people get a COVID vaccine in the hospital room |

#### A. GloVe

Global Vectors or GloVe[71] is an unsupervised learning algorithm for acquiring word vector portrayals. This is accomplished by projecting words into a significant space where the distance between words is identified with semantic similarity. Training is performed on a collected worldwide word-word co-occurrence matrix from a corpus, with features portraying fascinating linear substructures of the word vector space.

#### B. LSTM

LSTMs or Long Short-Term Memory [72] networks are derived from the Recurrent Neural Networks (RNNs) that can process data sequences. While RNNs can learn the context, however, due to backpropagation, there is an issue of vanishing gradients or exploding gradients, which LSTM can overcome. LSTM unit comprises multiple gates. These gates regulate the flow of information and allow some relevant information to pass through the sequence. So, this way, the model does not forget the information it has learned in the beginning. Therefore, 128 units of LSTMs were used in the architecture.

#### C. DANN

Domain Adversarial Neural Network (DANN) is a representation learning approach where the training and testing data sets are similar but belong to different distributions. The main inspiration behind the working of DANN is that for effective domain transfer to be achieved, predictions must be made based on overlapping features. Due to this, the network cannot discriminate between the training (source) and testing (target) domains[46]. Consequently, it does not require a labeled target domain dataset when using DANN but requires a labeled source domain data set. The DANN architecture consists of mainly three parts – feature extractor, label predictor, and domain classifier. The feature extractor consists of a deep neural network for performing feature learning. The label predictor and the feature extractor form a standard feed-forward neural architecture responsible for classifying the class label of the training data sample. Lastly, the domain classifier is accountable for achieving the unsupervised nature of DANN, in which it is connected to the feature extractor through the Gradient Reversal Layer (GRL). GRL has no parameters to be updated and acts as an identity transform during the forward propagation. Backpropagation multiplies the gradient obtained from the next layer by -1 and passes it to the previous layer, as mentioned in Eqn. 1. This layer makes sure that the feature distributions over the source and target domains are as similar as possible to obtain the domain invariant features. During training, those parameters of feature mapping are sought, which maximizes the loss of the domain classifier by making the source and target distributions as similar as possible and simultaneously minimizing the loss of the label predictor. DANN focuses on learning features that combine discriminativeness and domain invariance by optimizing the underlying features jointly.

$$\theta_f = \theta_f - \mu \left( \frac{dL_y}{d\theta_y} + (-1)(\lambda) \frac{dL_d}{d\theta_d} \right) \quad (1)$$

In Eqn 1, $\Theta f$ is the gradient of the feature extraction layer, $\mu$ is the learning rate, $\lambda$ is a hyperparameter for gradient reversal, Ly is label predictor loss, Ld is domain predictor loss, $\Theta y$ is parameters of label predictor, and $\Theta d$ is the parameters of domain predictor. This equation represents the update of feature parameters during backpropagation using GRL.

## V. SYSTEM DESIGN

The data collected from different sources was passed through a preprocessing pipeline and augmented to make it model-ready. This data was then vectorized using Glove embeddings before feeding it to a feature extractor. The Feature Extractor (FE) block consisted of sequential Conv1d, Max Pool layers, followed by an LSTM and Dense layer. Finally, the procured vectors from the feature extractor block were parallelly passed into Label Predictor and Domain Classifier (DC) blocks.

The Label Predictor (LP) was used to classify either of the labels, "True" or "False," using a Dense and a Sigmoid layer. The domain classifier also consisted of a Dense and a Sigmoid layer that predicted the input data's domain (source/target). During backpropagation, the gradient of the DC block would pass through a gradient reversal layer, as explained in section 4.1. Fig 2. gives a visual representation of the model architecture explained here.

For the misinformation detection application, we have used the Local Interpretable Model agnostic Explanation method to have a peek inside the black box DANN model to derive the target label prediction explainability of the DANN model as LIME is a post hoc model agnostic and can be applied to any classifier without change in model intrinsic and provide instance level local explanations on different multiple source modalities such as image, text and tabular data. Lime trains a linear model to approximate the local decision boundary of a particular instance. The technique can be applied to any classifier with text image and structured data providing a high degree of flexibility compared to the other methods giving it a significant weightage over others. The explanations are post hoc and are derived by approximating the underlying black box model locally by a linear interpretable model such as random forest [73]. This generates sparse explanations that are not too long and human-understandable. LIME is trained on small perturbations of original instances as local approximations around the predictions by sampling and obtaining a surrogate dataset for the input instance whose decision is to be explained. The weighting of features is based on how close they are to the original instance and the top significant features influencing the predictions are extracted. A random forest surrogate model with 500 trees is used for LIME implementation.

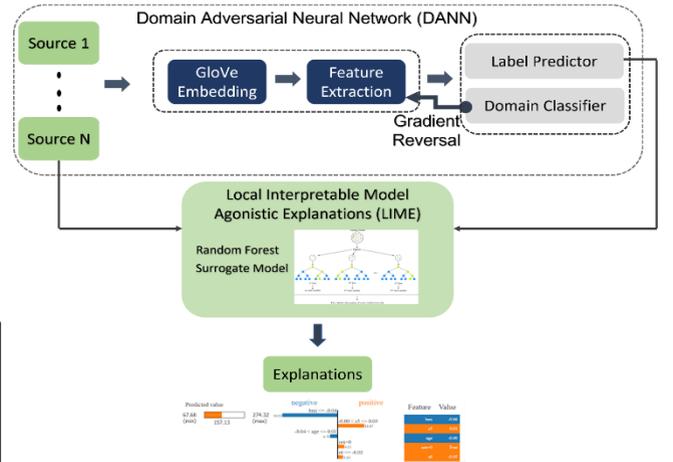

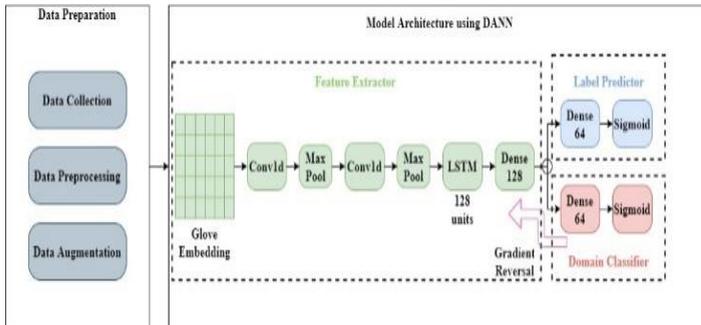

**FIGURE2.** Model Architecture for DANN

### A. Explainable Post hoc model using LIME

The different explainability methods for text classification in NLP include gradient-based saliency, integrated gradients IG, layerwise relevance propagation LRP and perturbation methods, feature importance-based techniques such as Shap, and attention-based heat maps showing model attention on specific words at a particular instance.

**FIGURE.3.** An integrated framework consisting of DANN and LIME for Explainable Misinformation Detection

### A. Model Evaluation and Comparison Procedure

A comprehensive testing methodology was implemented to test the effectiveness of the Domain adaptation approach. The first method, where the source dataset training was done using a FE+LP model and target data, was directly used as a testing component only. This would be referred to as "Without DANN" in the paper. Method 2 (With DANN) consisted of using DANN (FE+(LP, DC)) on the source and target data. Comparing "Without DANN' and "With DANN" approaches would help us demonstrate the approach's effectiveness over normal unsupervised techniques. News data comprised of the source domain and target domain would be the social media platforms like Twitter, Reddit, Instagram, etc. Metrics like Accuracy, Precision, Recall, and F1 score are used for evaluation. Additionally, Area Under the Curve of Receiver Operating Characteristic (AUC) is used to understand how well the model ranks correct and incorrect pairs. AUC tells us mathematically how the true positives rate grows at various false-positive rates.

## VI. RESULT AND DISCUSSION

Table 4 shows the different metrics of both the approaches, i.e., Without DANN and With DANN on the CoAID[68] dataset. It is observed that the DANN model performs better on the target domain (Twitter) when compared to the model trained on the source (News) and tested directly on the target (Twitter).

TABLE IV
RESULT OF COAID DATASET USING OUR METHOD

| Domain | Without DANN | | | With DANN | | |
|---|---|---|---|---|---|---|
| | Precision | Recall | F1 | Precision | Recall | F1 |
| News (Source) | 0.8336 | 0.8264 | 0.8296 | 0.7760 | 0.7805 | 0.7750 |
| Twitter (Target) | 0.9382 | 0.5212 | 0.6497 | 0.9502 | 0.7285 | 0.8113 |

Table 5. compares the results obtained by [68] to the results obtained using the DANN architecture mentioned in section 4. Precision, Recall, and F1 are some of the parameters for comparison [68]. DANN improves the Precision by 6% and the F1 score by 40% on target data. Here it is observed that DANN outperforms the previous approaches, and hence DANN can be used to learn features from one social media domain to another effectively.

TABLE V
RESULT ON COAID DATASET USING OTHER METHODS WITHOUT DANN

| Method | Precision | Recall | F1 |
|---|---|---|---|
| HAN [74] | 0.6965 | 0.4659 | 0.5471 |
| dEFEND [75] | 0.8965 | 0.4847 | 0.5814 |
| Our model (without DANN using FE + LP) on total CoAID dataset | 0.9701 | 0.9145 | 0.9300 |

Further, we present the results obtained on the MiSoVac data set in table 6. Accuracy/ AUC with DANN surpasses most of the social media platforms. In the MiSoVac dataset, the corpus for Twitter was balanced and substantial, so we observed a significant deviation for both the metrics in DANN and Without DANN. Using DANN, the accuracy of Twitter data increases by 22%, while AUC increases by 15% compared to the normal approach Without DANN. While the data for misinformation on other platforms was limited, we see only a slight change observed for the DANN approaches. Only Instagram shows a decrease in the accuracy by 10%; however, the AUC increases by 14% when DANN is applied, so the DANN model can distinguish the classes better even if the accuracy is low. An increase of 3% in the accuracy and 9% in the AUC scores are observed. We observe that DANN can learn the domain invariant features and is good at generalizing data from various social media platforms.

TABLE VI
RESULT OF MISOVAC DATASET WITH AND WITHOUT DANN

| Domain | Without DANN | | With DANN | |
|---|---|---|---|---|
| | Accuracy | AUC | Accuracy | AUC |
| News (Source) | 0.6393 | 0.6872 | 0.6885 | 0.8149 |
| Twitter (Target) | 0.6224 | 0.7413 | 0.7608 | 0.8493 |
| Instagram (Target) | 0.6945 | 0.6127 | 0.6247 | 0.7019 |
| Reddit (Target) | 0.6034 | 0.5214 | 0.6027 | 0.5610 |
| YouTube (Target) | 0.6250 | 0.6250 | 0.6250 | 0.6250 |

In table 7, the results were obtained by comparing the HAN[ Yang, 2016] architecture on the MiSoVac dataset alongside the "Without DANN" architecture. The "Without DANN" method performs better than the HAN[ Yang, 2016] architecture. Among the previous approaches mentioned, dEFEND [Shu, 2019] utilized a combination of tweets and their replies, so implementing it on the MiSoVac dataset was impossible.

TABLE VII
COMPARISON OF HANN WITH OUR METHOD WITHOUT DANN ON MISOVAC DATABASE

| Method | Precision | Recall | F1 |
|---|---|---|---|
| HAN [74] (on entire dataset) | 0.75 | 0.74 | 0.74 |
| Our model (without DANN using FE + LP) on total MiSoVAC dataset | 0.77 | 0.77 | 0.76 |

Below in figure 3 are plots that compare the accuracy and AUC of DANN and Without DANN implementation on the four target domains considered: Reddit, Instagram, Twitter, and YouTube.

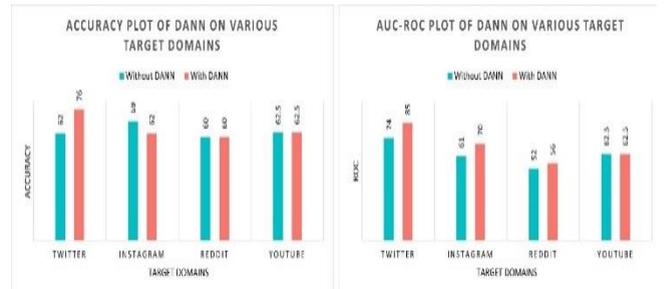

**FIGURE 4..Accuracy and AUC plot for various Social Media Platforms**

We generalize social media by combining multiple social media datasets into the source and target domains. We include samples from the News and Instagram datasets in the source domain, and the target domain has samples from Twitter, Reddit, and YouTube datasets. We perform Domain adaptive training using the DANN architecture mentioned in the previous section. These results are summarized in Table 8. When training the DANN model, the source accuracies (for combined news and Instagram) were 0.8, and the combined AUC was 0.9233. So, while training the DANN model, the model learned to understand complex features from data of various social media domains that increase the source results. As shown in Table 8, we see an improvement in accuracy and AUC for YouTube data for target results. There is a slight increase in the AUC of Twitter data while the accuracy is reduced. For reddit, as the number of test samples was relatively less, we see not much improvement in the results. The same was observed in the results mentioned in Table 7.

TABLE VIII
TESTING RESULTS FOR THE COMBINED AND INDIVIDUAL TARGET DOMAIN

| Media (Target Domain) | Without DANN | | With DANN | |
|---|---|---|---|---|
| | Accuracy | AUC | Accuracy | AUC |
| **Twitter** | 0.6945 | 0.6127 | 0.6562 | 0.7187 |
| **Reddit** | 0.6034 | 0.5214 | 0.5000 | 0.4800 |
| **YouTube** | 0.6250 | 0.6250 | 0.8999 | 0.8000 |
| **Combined (Twitter + Reddit + YouTube)** | 0.8846 | 0.9689 | 0.6730 | 0.6938 |

Explainable Model:

Explainability leads to disentanglement in the domain-specific features and improved generalization to the target domain without hindering performance on the source domain bridging the domain gap [76]. Domain shifts were in the data distribution of the source, and the target domain is different and can be addressed with XAI. Domain adaptation results in learning more discriminative features in the text classification results with the change in evidence in contrast to without domain adaptation, improving generalization on unseen domain learning domain invariant representation.

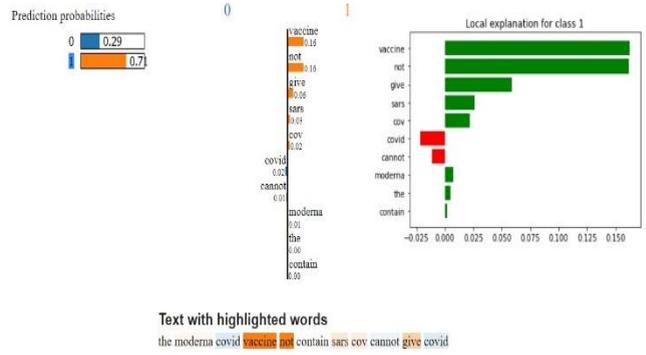

**FIGURE5.** Example – 1 LIME explanations for prediction probabilities for both classes (0 and 1) based on the score assigned to each word in the sentence text and its corresponding highlight color

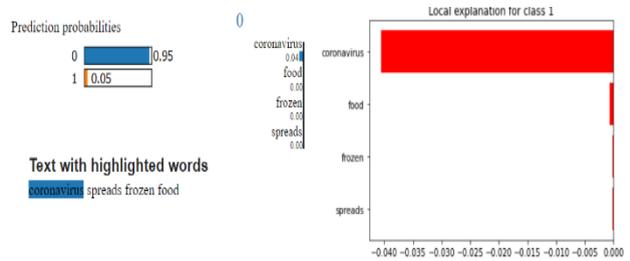

**FIGURE.5.** Example – 2 LIME explanations for prediction probabilities for both classes (0 and 1) based on the score assigned to each word in the sentence text and its corresponding highlight color. Only the class 0 word (coronavirus) is from the feature space.

.

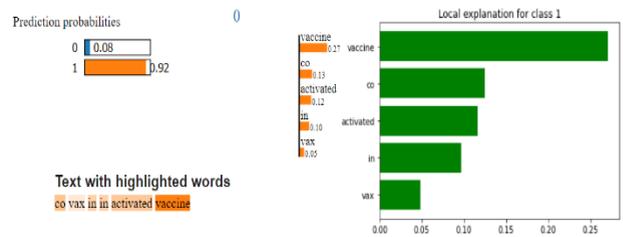

**FIGURE.6.** Example – 3 LIME explanations for prediction probabilities for both classes (0 and 1) based on the score assigned to each word in the sentence text and its corresponding highlight color. Only the class 0 word (coronavirus) is from the feature space

The LIME visualizations are intuitive and understandable. The results obtained are interpretable for humans. Local explanation reflects the local fidelity, i.e., the classifier's behavior for a particular data instance. Figures 4-6 demonstrate the output of the LIME for three class instances for the dataset.

The blue color represents class 0, and the orange color represents class 1. The text is highlighted with the probability of each word being in either class. The bar chart on the left shows float point numbers on the horizontal bars representing the relative importance of these features (green for class 1 and red for class 0). LIME maintains the explanatory ability of significant features regardless of the chosen classifier running independently of the model used. The text explainer finds the top words which primarily drive the model to make the classification decision providing intuitive model behavior. This maps with the original class label providing individual feature relevance and high feature contribution in the final prediction by highlighting the text. For Fig 4, the text has features with a higher probability (~70%) of being in class 1. In Fig. 6, a higher feature weight is given to the term vaccine with 92% class probability. In Fig. 5, the highest importance is assigned to the term coronavirus with the probability of 95% of class 0. LIME provides excellent results for text classifiers, but random sampling for data instances can be unstable in certain scenarios. It typically acts as an explanation tool for expert and non-expert users with diverse explainability requirements boosting trust for real-world adoption and deployment. The explanations are faithful and model agnostic. The model can distinguish between the actual and false sentences for misinformation detection.

## VII. CONCLUSION

In this paper, we demonstrated the explainable misinformation detection for social media platforms by employing the DANN and explainable AI LIME-based approach for explaining the target label predictions of the black box DAAN model making it more locally interpretable, trustworthy and adaptable in real-world applications. Persistent propagation of misinformation in the healthcare domain leads to a direct and significant impact on human social wellbeing; hence the adoption of explainable AI in establishing human trust is of paramount importance in the critical domain of healthcare. DANN is employed for misinformation detection across multiple social media platforms. We consider the most relevant case study in current times, namely COVID-19 misinformation, to implement and test our approach. We use two specific data sets, namely, CoAID, which is available openly and contains samples from news and Twitter sources. In addition to this, we developed a novel dataset named MiSoVac focusing on COVID-19 vaccine-related misinformation from various social media platforms. We described our data collection procedure, annotation, preprocessing techniques, and the architecture implemented to develop a generic classifier using DANN. Our methodology demonstrates promising results and outperforms other CoAID datasets' target domain approaches. An increase of ~40% was attained in the F1 score compared to the best model mentioned in the preexisting work [50]. We illustrate the effectiveness of DANN architecture on the MiSoVac dataset and observe that DANN surpasses results obtained by the Without DANN approach by ~5% in accuracy and ~11% in AUC on average across all target domains. Domain adaptation and explainability for various social platforms are still not explored extensively. We believe this is the first of many steps towards developing techniques capable of generalizing on multiple data of a similar domain. Our approach could prove more economical in terms of time and processing and generate significantly effective results without training the models for individual platforms powered with the joint approach of prediction and explanation for establishing trust and adoption. This work is not limited to misinformation detection and can be explored for more varied tasks in natural language classification, where the incoming data is an amalgamation of multiple domains. We also hope the MiSoVac dataset is helpful to fellow researchers to help tackle the COVID-19 misinformation spread.